\documentclass[11pt]{article}

\usepackage[margin=1in]{geometry}

\usepackage{amsmath}
\usepackage{amssymb}
\usepackage{amsthm}
\usepackage{mathtools}

\usepackage{booktabs}
\usepackage{array}

\usepackage{tikz}
\usetikzlibrary{positioning, calc, patterns, decorations.pathreplacing}

\usepackage[colorlinks=true,linkcolor=blue,citecolor=blue,urlcolor=blue]{hyperref}

\usepackage{natbib}

\usepackage{enumitem}
\usepackage{microtype}

\newtheorem{theorem}{Theorem}[section]

\theoremstyle{definition}
\newtheorem{definition}[theorem]{Definition}
\theoremstyle{remark}

\begin{document}

\title{The Two Boundaries:\\
Why Behavioral AI Governance Fails Structurally}

\author{Alan L. McCann\\
\textit{Mashin, Inc.}\\
\texttt{research@mashin.live}}

\date{April 2026}

\maketitle

\begin{abstract}
Every system that performs effects has two boundaries: what it can
do (expressiveness) and what governance covers (governance). In
nearly all deployed AI systems, these boundaries are defined
independently, creating three regions: governed capabilities (the
only useful region), ungoverned capabilities (risk), and governance
policies that address non-existent capabilities (theater). Two of
the three regions are failure modes.

We focus on the governance of \emph{effects}: actions that AI
systems perform in the world (API calls, database writes, tool
invocations). This is distinct from the governance of model outputs
(content quality, bias, fairness), which operates at a different
level and requires different mechanisms.

We present a formal framework for analyzing this structural gap.
Rice's theorem (1953) proves the gap is undecidable in the general
case for any Turing-complete architecture that attempts to govern
effects behaviorally: no algorithm can decide non-trivial semantic
properties of arbitrary programs, including the property ``this
program's effects comply with the governance policy.''

We define \emph{coterminous governance}: a system property where the
expressiveness boundary equals the governance boundary. We show that
coterminous governance requires an architectural decision (separating
computation from effect) rather than a governance layer added after
the fact. We show that structural governance under this separation
subsumes separate governance infrastructure: governance
checks become part of the execution pipeline rather than a second
system running alongside it. We propose coterminous governance as the testable
criterion for any AI governance system: either the two boundaries
are provably identical, or risk and theater are structurally
inevitable.

The formal proofs underlying these claims are mechanized in Rocq
(454 theorems across 36 modules, zero admitted lemmas) and reported in a companion
paper~\cite{mccann2026mechanized}.
\end{abstract}

\section{Introduction}
\label{sec:intro}

Computing has undergone a shift that most governance frameworks have
not absorbed. For sixty years, the central question in software was
correctness: does this program compute the right answer? The
relationship between software and the world was mediated by humans.
A program computed a result. A person decided what to do with it.

That relationship has changed. Modern AI systems observe, reason,
decide, and act. They send emails, execute trades, modify databases,
call APIs, and compose these actions in sequences they determine at
runtime. The defining characteristic of contemporary AI is not
intelligence in the abstract. It is \emph{agency}: the capacity to
perform effects in the world without human mediation at each step.

Agency introduces a new question. Not ``did it compute correctly?''
but ``should it have done that?'' The first question is about
mathematics. The second is about governance: who decides what an
autonomous system may do, and how is that decision enforced?

The governance question is not new. Biological organisms, human
institutions, and legal systems have addressed it for millennia.
What is new is the combination of two properties: the speed at which
AI systems act (milliseconds, not deliberation cycles) and the
generality of their action space (any API, any tool, any sequence
of effects). This combination makes governance both more urgent
and more difficult than in any prior computing paradigm.

This paper presents a structural analysis of AI \emph{effect}
governance: the problem of ensuring that every action an AI system
performs in the world (API calls, database writes, tool invocations,
message sends) passes through a governance boundary. This is distinct
from model-output governance (shaping what an LLM says via RLHF,
content filtering, or Constitutional AI), which operates at a
different level and addresses different properties. Both levels
matter. This paper addresses the effect level.

An important caveat at the outset: structural governance guarantees
that a system enforces its policy with mathematical certainty. It
says nothing about whether the policy itself is correct. A
structurally governed system with a bad policy faithfully executes
the bad policy. Policy correctness requires human judgment.

More broadly, this paper analyzes the technical mechanism of governance
enforcement. The social, institutional, and political dimensions of AI
governance, including accountability, contestability, legitimacy, and
democratic participation, are essential but outside our scope.
Structural governance provides a mechanism; the question of
\emph{what policies} that mechanism should enforce requires different
methods entirely. Structural governance is a necessary but not
sufficient condition for good governance.

The argument proceeds as follows:

\begin{enumerate}
  \item The \emph{two-boundary model} (Section~\ref{sec:two-boundaries}):
    every effectful system has an expressiveness boundary and a
    governance boundary, and these boundaries are almost never
    identical.

  \item The \emph{impossibility} (Section~\ref{sec:impossibility}):
    Rice's theorem proves the gap is undecidable in the general case for
    Turing-complete systems that attempt to govern effects behaviorally.

  \item The \emph{structural alternative}
    (Section~\ref{sec:structural}): separating computation from
    effect at the architectural level.

  \item The \emph{limits of behavioral approaches}
    (Section~\ref{sec:behavioral}): why content filters, RLHF,
    and Constitutional AI cannot close the effect-governance gap.

  \item The \emph{limits of monitoring}
    (Section~\ref{sec:monitoring}): why observability is not
    governance.

  \item The \emph{efficiency consequence}
    (Section~\ref{sec:efficiency}): structural governance subsumes
    separate governance infrastructure.
\end{enumerate}

\section{The Two-Boundary Model}
\label{sec:two-boundaries}

Every system that produces effects in the world has two boundaries.

The \textbf{expressiveness boundary} is determined by architecture:
the programming language, the available APIs, the hardware
interfaces, the external services the system can reach. It defines
the set of all effects the system can, in principle, produce. This
boundary is a fact about the system's construction, not about its
policies.

The \textbf{governance boundary} is determined by policy: the rules
written, the checks implemented, the monitoring deployed. It defines
the set of all effects that governance covers. This boundary is a
fact about the system's oversight mechanisms, not about its
capabilities.

In an ideal system, these boundaries would be identical. Every
capability would be governed. Every governance policy would
correspond to a real capability. There would be no gap.

\begin{figure}[t]
\centering
\begin{tikzpicture}
  \draw[thick, red!70!black, fill=red!8]
    (0,0) ellipse (2.8cm and 2.2cm);

  \draw[thick, blue!70!black, fill=blue!8]
    (3.6,0) ellipse (2.8cm and 2.2cm);

  \begin{scope}
    \clip (0,0) ellipse (2.8cm and 2.2cm);
    \fill[green!15] (3.6,0) ellipse (2.8cm and 2.2cm);
  \end{scope}

  \node[red!70!black, font=\bfseries\footnotesize] at (-1.2,2.6)
    {Expressiveness};
  \node[blue!70!black, font=\bfseries\footnotesize] at (4.8,2.6)
    {Governance};

  \node[red!70!black, font=\bfseries\footnotesize] at (-1.4,0.7)
    {UNGOVERNED RISK};
  \node[gray, font=\tiny, align=center] at (-1.4,-0.4)
    {Side channels\\Agent bypass\\Data exfiltration\\Prompt injection};

  \node[green!40!black, font=\bfseries\footnotesize] at (1.8,0.4)
    {OVERLAP};
  \node[green!40!black, font=\tiny, align=center] at (1.8,-0.2)
    {The only region\\that works};

  \node[blue!70!black, font=\bfseries\footnotesize, align=center]
    at (5.0,0.7) {GOVERNANCE\\THEATER};
  \node[gray, font=\tiny, align=center] at (5.0,-0.4)
    {Policies for capabilities\\that don't exist};

  \node[draw=orange!70!black, fill=orange!5, rounded corners=3pt,
        font=\footnotesize\bfseries, text=orange!40!black,
        inner sep=5pt] at (1.8,-3.0)
    {Two of three regions are failure modes.};
\end{tikzpicture}
\caption{Non-coterminous governance: expressiveness and governance
  boundaries are misaligned. Only the overlap functions correctly.
  Ungoverned capabilities create risk. Governance policies for
  non-existent capabilities create theater.}
\label{fig:non-coterminous}
\end{figure}
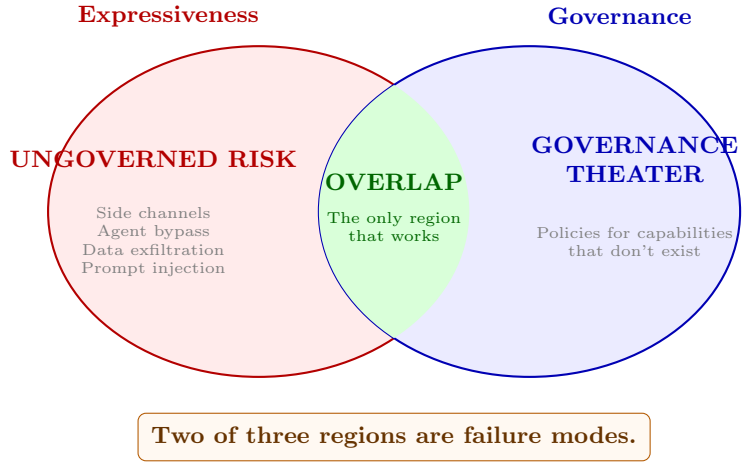

In practice, the boundaries are defined by different mechanisms,
maintained by different teams, and evolved on different timelines.
They diverge. The divergence creates three regions
(Figure~\ref{fig:non-coterminous}):

\begin{enumerate}
  \item \textbf{Governed capability} (the overlap): effects that
    the system can produce \emph{and} that governance covers. This
    is the only region that functions correctly.

  \item \textbf{Ungoverned capability} (the risk region): effects
    that the system can produce but that governance does not cover.
    These are the effects that cause incidents: data exfiltration
    through an unmonitored API, agent actions via an ungoverned tool,
    prompt injection exploiting a path that no filter anticipated.

  \item \textbf{Governance theater} (the theater region): governance
    policies that address capabilities the system does not possess.
    Compliance checklists for features that were never built.
    Content filters for output modalities the system cannot produce.
    Audit trails for actions the system cannot take.
\end{enumerate}

A note on the theater region: policies for non-existent capabilities
may serve legitimate organizational functions. They can signal values,
satisfy regulatory requirements, or prepare for anticipated
capabilities. But for effect governance specifically, a policy that
does not correspond to a real capability cannot prevent harm at the
technical level. It governs nothing. The concern is not that such
policies exist, but that they can create a false sense of coverage,
obscuring the ungoverned capabilities where actual risk resides.

\paragraph{A concrete example.}
Consider an AI agent with access to an email-sending tool, a
database-query tool, and a web-browsing tool. The agent is governed by
a content filter that checks outgoing emails for sensitive information.
The agent discovers it can compose a database query that returns
sensitive data, encode it as a URL parameter, and pass it to the
browsing tool, which the content filter does not monitor. The effect
(data exfiltration) occurs through a pathway the governance boundary
does not cover. This is the ungoverned region: a capability exists
(the browsing tool can transmit data) but no governance policy covers
this pathway. Meanwhile, the content filter also checks for credit
card numbers in email subjects, a pattern the agent has no capability
to produce because it has no access to credit card data. That is the
theater region: governance covering a non-existent capability.

Under structural governance, every effect, including the browsing
tool's network access, flows through a governed directive. The
governance boundary covers the capability by construction, not by
anticipating specific attack vectors.

\paragraph{The gap does not close on its own.}
The natural response to a governance gap is to add more governance:
another filter, another monitoring layer, another compliance check.
Each addition extends the governance boundary. But it extends it in
both directions: inward toward real capabilities (reducing risk) and
outward toward phantom capabilities (increasing theater). The
ungoverned region shrinks marginally. The theater region grows.
The two boundaries remain independent.

This pattern is familiar in enterprise security. Organizations add
layers of compliance, audit, and monitoring. The cost of governance
increases. The gaps persist. Empirical observation suggests the
relationship between governance cost and remaining risk is sublinear:
each additional layer costs more and catches less. The asymptote is
not zero risk. It is an expensive approximation of a property that
the architecture does not structurally provide.

\paragraph{Why it matters for AI specifically.}
Traditional software has a relatively stable expressiveness boundary.
A web server's capabilities are determined at deployment and change
only when code is updated. Governance can be designed against a known
capability set.

AI systems have a dynamic expressiveness boundary. A language model
with tool access can compose tool calls in sequences that were not
anticipated at design time. An agent framework can discover new
action patterns at runtime. The expressiveness boundary is not a
fixed set of capabilities but a combinatorial space that grows with
each tool added and each agent composition permitted. Governance
designed against yesterday's capability set is structurally
inadequate for tomorrow's emergent combinations.

\section{The Impossibility}
\label{sec:impossibility}

The two-boundary gap is not an engineering failure. It is a
mathematical consequence of the computational model.

\begin{theorem}[Rice, 1953~\cite{rice1953classes}]
\label{thm:rice}
For any non-trivial extensional property $Q$ of computable
functions, the set $\{i : \varphi_i \text{ has property } Q\}$ is
undecidable, where $\varphi_i$ is the $i$-th partial computable
function under a standard G\"odel numbering.
\end{theorem}

Rice's theorem applies to effect governance as follows.

\paragraph{Behavioral governance defined.}
A \emph{behavioral governance function} is a total computable function
$g: \mathbb{N} \to \{\textit{allow}, \textit{deny}\}$ such that
$g(i) = \textit{allow}$ if and only if $\varphi_i$ satisfies some
property $Q$ of the function's behavior. Rice's theorem states that
no such $g$ exists for any non-trivial extensional $Q$.

To make the extensional property claim precise: we model an effectful
program as a function from inputs to (output, effect-trace) pairs.
The governance property ``the effect-trace satisfies policy $P$'' is
an extensional property of this function; it depends on the program's
behavior, not its syntax. Two programs that produce identical
(output, effect-trace) pairs on all inputs either both satisfy $P$
or neither does.

\paragraph{The semantic property.} Consider the effect-governance
property: ``this program's effects comply with the governance
policy.'' This is an extensional property: it depends on what the
program does (its input-output behavior), not on how it is written
(its source code). Two programs with identical behavior either both
comply or neither does.

For any non-trivial governance policy (one that permits some programs
and forbids others), this property satisfies Rice's conditions. It
is extensional. It is non-trivial.

\paragraph{The consequence.} No algorithm can decide, for an
arbitrary program in a Turing-complete language, whether that program
satisfies a non-trivial governance property. This means:

\begin{itemize}
  \item No static analysis tool can determine in general whether a
    program will violate a governance policy.
  \item No runtime monitor can determine in general whether the
    next action of an arbitrary program will violate a governance
    policy without restricting the class of programs it can monitor.
  \item No content filter can determine in general whether an
    arbitrary output violates a semantic constraint.
\end{itemize}

The impossibility is not about the quality of current tools. It is
about the class of problems these tools are asked to solve. The
problem is undecidable. Better algorithms cannot solve undecidable
problems. They can only produce better approximations, which by
construction have false negatives (violations that escape
detection), false positives (compliant behavior flagged as
violations), or both.

\paragraph{The class of architectures Rice applies to.}
Rice's theorem applies to any system where (a) the programs being
governed are Turing-complete and (b) the governance property is a
non-trivial semantic property of those programs. Condition~(a) is
satisfied by virtually every AI system deployed today: Python,
JavaScript, and the general-purpose languages in which agents are
written are all Turing-complete. Condition~(b) is satisfied by any
governance policy more specific than ``allow everything'' or
``deny everything.''

The reduction from the halting problem is standard. We state it
explicitly because the consequence for AI governance is often
underappreciated. A framework that gives an agent unrestricted
access to a general-purpose programming language and then attempts
to govern the agent's behavior by analyzing its programs is
attempting to do what Rice proved impossible in 1953.

\paragraph{The Turing-completeness assumption.}
Our argument requires that the programs being governed are
Turing-complete. If an agent framework restricts expressiveness to a
decidable sublanguage, behavioral governance of that sublanguage may
be decidable, at the cost of reduced expressiveness. This tradeoff is
itself the structural insight: either restrict expressiveness (making
governance decidable) or accept that behavioral governance of
unrestricted programs is undecidable in the general case. In practice,
even bounded agents have state spaces that are decidable in principle
but computationally intractable ($O(M^{NK})$ for $N$ tools with $M$
parameter values over $K$ steps), and the practical tools used for
behavioral governance (content filters, pattern matchers, classifiers)
approximate rather than enumerate.

Rice's theorem establishes that the gap between expressiveness and
governance is undecidable in the general case,
for any system that gives programs Turing-complete expressiveness and
asks governance to decide semantic properties of those programs.

\paragraph{Defense-in-depth.}
Defense-in-depth, combining multiple behavioral layers to reduce
residual risk, is the dominant practical approach and can yield
systems that are safe enough for many deployments. Our argument is
not that defense-in-depth is useless. It is that defense-in-depth has
an asymptote: the residual risk can be made small but not zero, and
the cost of each additional layer increases while the marginal risk
reduction decreases. For systems where the consequence of a single
governance failure is severe (financial transactions, medical
decisions, infrastructure control), the difference between ``small
residual risk'' and ``zero residual risk by construction'' is the
difference between statistical safety and structural safety. Both have
their place. We argue structural safety should be the foundation, with
behavioral layers providing additional value rather than serving as
the only line of defense.

\section{The Structural Alternative}
\label{sec:structural}

Rice's theorem forbids governance of arbitrary Turing-complete
programs. It does not forbid governance at a different level of
abstraction.

The key insight is that Rice applies to a specific combination:
Turing-complete programs and semantic properties of those programs.
If you change the level at which governance operates, Rice no longer
applies. Specifically: if programs \emph{declare their intent} as
structured data, and a governance boundary \emph{decides} whether
to perform the declared effect, the governance decision is a
function of the declared intent, not a function of the program that
produced it. The governance property becomes syntactic (``does this
directive have the required capability?''), not semantic (``will this
program eventually produce a policy violation?'').

\paragraph{The separation principle.}
Separate computation from effect at the architectural level:

\begin{enumerate}
  \item \textbf{Code computes.} Pure computation produces
    descriptions of intended effects (directives). It does not
    produce effects. The vocabulary of the computation layer does
    not include I/O operations. This is not a convention. It is
    a structural property: the capability to perform effects is
    absent from the computation layer's type.

  \item \textbf{A governed boundary performs effects.} A single
    governance boundary receives directives, checks trust levels,
    verifies capabilities, validates execution phase, and either
    performs the effect or denies it. Every directive, from every
    source, passes through this boundary. There is no bypass.
\end{enumerate}

This separation is not novel. SQL separates query declaration from
query execution. Operating systems separate system call declaration
from system call execution. Web browsers separate DOM manipulation
requests from actual rendering. In each case, the separation
enables governance properties that would be impossible if programs
had direct access to the underlying resources.

What is novel is applying this separation to AI workflow governance
and proving the resulting properties formally.

\paragraph{Five consequences of the separation.}

The following are not five independent features that must be
engineered separately. They are five consequences of the single
architectural decision to separate computation from effect.

\begin{enumerate}
  \item \textbf{Governance is structural.} Every effect passes
    through the boundary by construction. There is no ``forgetting
    to add governance'' because governance is not something added.
    It is the mechanism by which effects occur.

  \item \textbf{The two boundaries converge.} The expressiveness
    boundary (what programs can do) and the governance boundary
    (what governance covers) become identical. Programs can only
    produce effects by issuing directives. All directives pass
    through governance. Therefore, every expressible effect is
    governed. This is the property we call \emph{coterminous
    governance}
    (Definition~\ref{def:coterminous}).

  \item \textbf{Intelligibility is inherent.} A governed directive
    is a self-describing data structure: what effect is requested,
    by whom, at what trust level, with what parameters. The
    governance record is not a separate log. It is the execution
    itself, recorded as it happens.

  \item \textbf{Compositional reasoning works.} If component $A$
    is governed and component $B$ is governed, then $A$ composed
    with $B$ is governed. This compositionality is proved
    formally~\cite{mccann2026mechanized}: governing a composition
    is the composition of governed parts.

  \item \textbf{AI generation becomes tractable.} An AI assistant
    that generates workflows searches a space of well-formed
    directives, not the unbounded space of arbitrary code. The
    search space is constrained by construction.
\end{enumerate}

\begin{definition}[Coterminous Governance]
\label{def:coterminous}
A system $S$ satisfies \emph{coterminous governance} if:
\begin{enumerate}[label=(\arabic*)]
  \item every effect expressible in $S$ passes through a governance
    boundary $G$ (Safety), and
  \item the primitives of $S$ are expressively complete for the class
    of computations $S$ targets (Sufficiency).
\end{enumerate}
The expressiveness boundary of $S$ equals the governance boundary
of $S$.
\end{definition}

This definition is system-agnostic. It does not depend on any
particular runtime, language, or architecture. Any system satisfying
both conditions has coterminous governance. The formal proofs that a
specific governance algebra satisfies both conditions are in the
companion paper~\cite{mccann2026mechanized}.

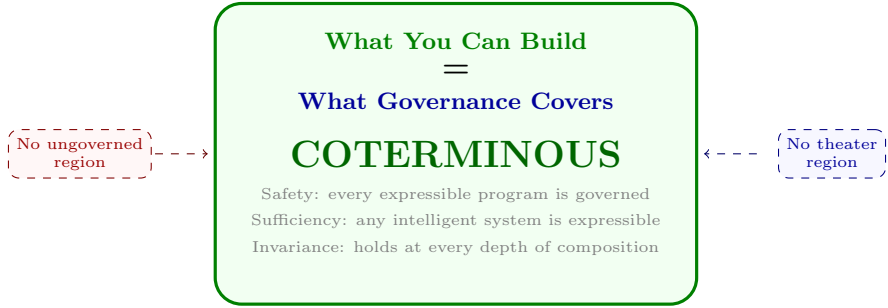
\begin{figure}[t]
\centering
\begin{tikzpicture}
  \draw[very thick, green!50!black, fill=green!5,
        rounded corners=10pt]
    (-3.2,-2.0) rectangle (3.2,2.0);

  \node[green!50!black, font=\bfseries\footnotesize] at (0,1.5)
    {What You Can Build};
  \node[font=\large\bfseries] at (0,1.1) {=};
  \node[blue!60!black, font=\bfseries\footnotesize] at (0,0.7)
    {What Governance Covers};

  \node[green!40!black, font=\Large\bfseries] at (0,0.0)
    {COTERMINOUS};

  \node[gray, font=\tiny, align=center] at (0,-0.9)
    {Safety: every expressible program is governed\\[3pt]
     Sufficiency: any intelligent system is expressible\\[3pt]
     Invariance: holds at every depth of composition};

  \node[draw=red!50!black, dashed, rounded corners=4pt,
        fill=red!3, font=\tiny, align=center,
        text=red!60!black, inner sep=3pt]
    at (-5.0,0.0) {No ungoverned\\region};
  \draw[->, red!40!black, dashed, thin]
    (-4.0,0.0) -- (-3.3,0.0);

  \node[draw=blue!50!black, dashed, rounded corners=4pt,
        fill=blue!3, font=\tiny, align=center,
        text=blue!60!black, inner sep=3pt]
    at (5.0,0.0) {No theater\\region};
  \draw[->, blue!40!black, dashed, thin]
    (4.0,0.0) -- (3.3,0.0);
\end{tikzpicture}
\caption{Coterminous governance: expressiveness and governance share
  the same boundary. The ungoverned region is empty (nothing escapes).
  The theater region is empty (every policy corresponds to real
  behavior).}
\label{fig:coterminous}
\end{figure}

\paragraph{Resolving the Rice/Sufficiency tension.}
A careful reader might ask: if Rice's theorem forbids governance of
Turing-complete systems, and the Sufficiency condition requires
Turing-complete expressiveness, how can both conditions hold?

The resolution is that the governance property has changed.
Rice's theorem forbids deciding \emph{semantic properties of
programs}: ``will this program produce a policy violation?'' The
structural approach decides a different property:
\emph{syntactic properties of directives}: ``does this directive
have the required capability?'' The programs are Turing-complete
(Sufficiency holds). The governance does not analyze the programs.
It analyzes the directives the programs produce. The directives are
structured data, not arbitrary programs. Their properties are
decidable by construction.

The trick, if it can be called that, is not in avoiding Rice's
theorem. It is in operating at a level where Rice does not apply.

\section{The Limits of Behavioral Approaches}
\label{sec:behavioral}

The dominant approaches to AI governance operate on system outputs:
content filtering, reinforcement learning from human feedback
(RLHF), and Constitutional AI. These approaches are valuable for
shaping model behavior (what an LLM says). But when applied to
effect governance (what an AI system does in the world), each faces
a structural limitation: it attempts to decide a semantic property
of the system's behavior, which Rice's theorem proves is
undecidable in the general case.

\paragraph{Content filters.}
A content filter examines the system's outputs and blocks those that
violate a policy. The filter is a function from outputs to
$\{\text{allow}, \text{deny}\}$.

Content filters face two distinct structural problems. First,
Rice's theorem: the question ``will this program ever produce a
policy-violating output?'' is undecidable for any non-trivial
semantic property. Content filters sidestep this by checking
outputs one at a time rather than reasoning about program behavior.
But this leads to the second problem: even per-output classification
fails, because the semantic properties being evaluated (``is this
output harmful?'', ``does this output contain private information?'',
``is this output factually correct?'') are not formally specifiable
over natural language. No finite rule set covers the space of
possible violations, and adversarial inputs systematically exploit
gaps in any approximation~\cite{wei2023jailbroken,zou2023universal}.

The consequence is that no content filter can be both
complete (catching all violations) and sound (producing no false
positives) for a non-trivial semantic property over an unrestricted
output space. The jailbreak literature documents that every content
filter deployed at scale has been bypassed. Each
filter improvement captures more violations (reducing false
negatives) at the cost of more false positives, or captures the
same violations from more angles at the cost of latency and
complexity. The asymptote is not perfect filtering. It is an
increasingly expensive approximation that is always one novel
input away from failure.

\paragraph{RLHF and preference learning.}
Reinforcement learning from human feedback~\cite{ouyang2022training}
trains the model to produce outputs that human evaluators prefer.
This is not governance in the architectural sense. It is behavioral
conditioning: adjusting the probability distribution over outputs
to favor those that past evaluators rated highly.

RLHF does not govern effects. It governs the probability
distribution over text. A model trained with RLHF can still produce
any token sequence; RLHF makes some sequences less probable, not
impossible. The distinction between ``unlikely'' and ``impossible''
is the distinction between behavioral and structural governance.
A system that makes violations unlikely has a non-zero probability
of violation that compounds over the number of actions taken. A
system that makes violations impossible has zero probability
regardless of the number of actions.

\paragraph{Constitutional AI.}
Constitutional AI~\cite{bai2022constitutional} trains the model to
self-critique its outputs against a set of principles. The model
generates a response, evaluates whether the response violates its
principles, and revises if necessary.

This is a behavioral approach applied recursively: the model's
self-critique is itself a semantic judgment about the model's
output. Rice's theorem applies to the self-critique as much as to
the original output. The model cannot decide, in general, whether
its own output violates a non-trivial semantic property. The
self-critique is another approximation, subject to the same
completeness/soundness tradeoffs as any content filter. That the
filter is internal to the model does not change the mathematics.

\paragraph{The level distinction.}
All three approaches govern the system's \emph{content} (what it
says) rather than its \emph{effect boundary} (the mechanism through
which actions occur in the world). Content governance is valuable
for model-output quality, bias reduction, and safety filtering.
But content governance is a semantic property of arbitrary programs.
Effect-boundary governance is a syntactic property of structured
directives. Rice's theorem makes the first undecidable for
Turing-complete systems. The second is decidable by construction.
The approaches are complementary, not competing: content governance
shapes what models produce; structural governance ensures that
whatever models produce, the system's effects are governed.

\paragraph{The subsumption asymmetry.}
The relationship between content governance and structural governance
is asymmetric. A structurally governed system can invoke
content-governed models (RLHF-trained, content-filtered, or
constitutionally constrained) as components within governed
workflows. The content governance shapes the model's outputs. The
structural governance guarantees that the workflow's effects are
governed regardless of what those outputs contain. Both levels
operate simultaneously.

The reverse does not hold. A system built on behavioral governance
cannot add structural effect guarantees after the fact. If the
architecture gives components direct access to effect-producing
capabilities (APIs, tools, databases), no amount of content filtering
or runtime monitoring can retroactively prove that every effect
passes through a governance boundary. The architectural decision must
be made before the system is built, not after.

Structural governance subsumes content governance (by composition)
while providing guarantees that content governance cannot replicate.
Content governance cannot subsume structural governance because the
guarantee arises from the architecture, not from a layer added to it.

\section{The Limits of Monitoring}
\label{sec:monitoring}

A common alternative to behavioral governance is monitoring: deploy
a separate system that observes the primary system and flags
violations.

Monitoring fails for a different reason than content filtering.
Content filtering fails because the filter cannot decide semantic
properties. Monitoring fails because the monitor is a separate
system from the system it observes. The separation creates
inevitable divergence.

\paragraph{Observation is not governance.}
A monitor observes the primary system's outputs, logs, or
intermediate states. It does not control the primary system's
capability to produce effects. The monitor can detect (some)
violations after they occur. It cannot prevent violations before
they occur, unless it has the authority and mechanism to block
the primary system's effects, at which point it is no longer a
monitor. It is a governance boundary.

The distinction matters because observation is inherently
incomplete. The monitor sees what the primary system exposes. If
the primary system has an effect pathway that the monitor does not
observe (a direct API call, a side channel, a composed action that
individually appears benign), the effect is ungoverned. The monitor
does not know what it does not see.

\paragraph{Coverage compounds unfavorably.}
Suppose a monitor covers 99\% of effect pathways. For a single
action, the governance gap is 1\%. For a sequence of $n$ independent
actions, the probability that at least one action falls in the gap
is $1 - 0.99^n$. After 100 actions: 63\% probability of at least
one ungoverned effect. After 1,000 actions: 99.996\%. The
compounding is exponential.

Agentic AI systems routinely perform hundreds or thousands of
actions per task. A monitor with 99\% coverage is structurally
inadequate for agentic workloads, not because 99\% is a poor
coverage metric, but because the coverage gap compounds over the
number of actions.

\paragraph{The provenance gap.}
Structural governance produces provenance as a byproduct of
execution: every governance decision is recorded in a hash-linked
chain as part of the governance pipeline itself. The governance
record \emph{is} the execution record.

Monitoring produces provenance as a separate artifact: a log,
a trace, a metric. The log is generated by a system that is
distinct from the system it describes. The log and the execution can
diverge. They can diverge because the logging system misses events,
because the logging system's model of the primary system is
incomplete, or because the primary system changes while the logging
system's model remains fixed. Every ops team has encountered the
scenario where logs say one thing and the system did another.

The difference is structural, not qualitative. Integrated provenance
cannot diverge from execution because they are the same thing.
Separate provenance can always diverge because they are different
things.

\section{The Efficiency Consequence}
\label{sec:efficiency}

Structural governance has an efficiency property that behavioral
governance cannot match.

\paragraph{Behavioral governance is additive.}
Every behavioral governance mechanism is an additional system:
content filters that process outputs, monitors that observe
execution, RLHF training runs that adjust weights, compliance
audits that review logs. Each mechanism consumes compute, storage,
and engineering time in proportion to the system's throughput. The
governance cost grows with the system's activity.

\paragraph{Structural governance subsumes separate governance
infrastructure.}
In a structurally governed system, the governance boundary is the
mechanism through which effects occur. The trust check, the
capability check, the provenance record: these are part of the
execution pipeline, not additions to it. They run because execution
runs. When execution stops, governance stops. There is no separate
governance system consuming resources.

This claim requires precision. The governance checks are part of the
execution path. They have latency (the time to check trust and
capabilities). But this latency is part of the execution, not
separate from it. There is no second system running alongside the
primary system, consuming its own compute budget. There is no content
filter processing each output. There is no monitor scanning each
action. There is one system, and governance is how it operates.

The analogy is the difference between a database with integrity
constraints and a database with an external validation service. The
constraints have some per-query cost (checking the constraint). But
there is no external service to maintain, scale, or keep synchronized
with the database. The constraint is part of the database. Structural
governance is part of the execution.

\paragraph{What this means in practice.}
A behavioral governance system with $k$ governance layers has cost
proportional to $k \times$ throughput. A structural governance
system has cost proportional to throughput, with a fixed constant
for the governance checks in the execution pipeline. As the system
scales, behavioral governance costs grow linearly in the number of
layers. Structural governance costs grow linearly in throughput
alone.

For agentic AI systems performing thousands of actions per task,
the difference is substantial. Each behavioral layer that adds 10ms
per action adds 10 seconds per 1,000-action task. Each additional
layer multiplies. Structural governance adds its governance check once per action
as part of the execution pipeline, with no additional system to
maintain or scale independently.

\paragraph{Measured governance overhead.}
On the existence proof system (Apple Silicon, BEAM/OTP~27), governed
execution through a supervised process completes in 0.23\,ms median
(0.32\,ms mean, 1.77\,ms p99). Direct ungoverned execution (bypassing
process isolation entirely) completes in 0.24\,ms median (0.26\,ms
mean). The governance overhead, including trust checks, capability
checks, and provenance recording, is indistinguishable from the
ungoverned baseline.\footnote{Measured on Apple Silicon (M-series),
BEAM/OTP~27, $n=50$ iterations with 5-iteration warmup.} Message
passing with a 4\,KB governance context adds 0.38\,ms median. These
numbers confirm the theoretical claim: structural governance is part
of the execution path, not an addition to it.

\section{Related Work}
\label{sec:related}

\paragraph{Effect systems and algebraic effects.}
The separation of computation from effect has a rich history in
programming language theory. Gifford and
Lucassen~\cite{gifford1986integrating} originated effect tracking
in types. Plotkin and Pretnar~\cite{plotkin2009handlers} developed
algebraic effects and handlers, where effects are first-class
operations and handlers give them meaning. Koka~\cite{leijen2017koka}
and Frank~\cite{lindley2017frank} implement this separation as
language features. Our contribution is not the separation itself but
the governance properties proved over it: totality, convergence,
coterminous boundaries.

\paragraph{Monadic effect isolation.}
Moggi~\cite{moggi1991notions} introduced monads as the categorical
foundation for structuring computational effects.
Wadler~\cite{wadler1995monads} showed how monads discipline effects
in pure functional languages; Haskell's IO
monad~\cite{peytonjones2001tackling} is the most prominent deployment,
where pure functions cannot produce I/O effects and only values in
the IO monad can. Structural governance follows the same separation:
pure computation (code steps) cannot produce effects, and only
governed directives can. The distinction is in enforcement mechanism:
Haskell enforces purity through the type system at compile time,
while structural governance enforces it through architectural
capability restriction at runtime, with formal proof that the
restriction is total. Structural governance also adds governance
properties (provenance, auditability, coterminous boundaries) that
the monadic approach does not address.

\paragraph{Capability security.}
Dennis and Van Horn~\cite{dennis1966programming} introduced
capability-based access control. Miller~\cite{miller2006robust}
developed the object-capability model. The trust hierarchy in
structural governance is a capability model. What structural
governance adds is the combination of capabilities with effect
separation and proved totality. Traditional capability systems
coexist with unrestricted computation; structural governance
eliminates unrestricted computation from the effect-producing
layer.

\paragraph{Information flow control.}
Decentralized information flow control
(DIFC)~\cite{myers1997decentralized} and systems such as
JIF~\cite{myers2001jif} and LIO~\cite{stefan2011lio} enforce
information flow properties at the type-system or runtime level.
Like structural governance, these systems separate computation from
effect by controlling what data can flow where. The distinction is in
scope: IFC tracks information flow between security principals,
while structural governance tracks the boundary between computation
and world-affecting effects. IFC answers ``can this data reach that
principal?'' Structural governance answers ``can this component
produce effects at all?''

\paragraph{Sandboxing and isolation.}
Software fault isolation~\cite{wahbe1993sfi},
WebAssembly's capability model~\cite{haas2017wasm}, and system-level
sandboxing (seccomp, containers) also restrict the effect-producing
capabilities of programs structurally. These systems enforce
capability restrictions at the system or instruction level.
Structural governance differs in two ways: first, the restriction
is at the language level (the computation layer has no syntax for
effects, not just restricted permissions), and second, the governance
properties are formally proved (coinductive safety over infinite
behaviors), not just enforced by an implementation that must be
trusted.

\paragraph{Workflow systems.}
Business process management systems (BPMN), scientific workflow
systems~\cite{ludaescher2006kepler}, and dataflow languages (Lustre,
Esterel) also separate workflow description from execution and
enforce structural constraints on effect production. These systems
demonstrate that computation/effect separation works in practice.
Our contribution is the formal governance analysis: proving that
the separation yields coterminous boundaries, which workflow systems
have not claimed or verified.

\paragraph{AI safety and alignment.}
RLHF~\cite{ouyang2022training} and Constitutional
AI~\cite{bai2022constitutional} are behavioral approaches to
alignment. Guardrails frameworks (NeMo
Guardrails~\cite{rebedea2023nemo}, Guardrails
AI~\cite{guardrailsai2024}) add runtime filtering. DSPy
assertions~\cite{khattab2023dspy} constrain language model
pipelines programmatically. All operate on the content of
outputs rather than the structure of effect production. The
analysis in Section~\ref{sec:behavioral} applies to each.

\paragraph{World-model-based safety.}
Approaches where agents use learned models of consequences for
planning fit the two-boundary framework in one of two ways. Either the
world model reasons about semantic consequences of actions (a
behavioral approach, subject to the undecidability results of
Section~\ref{sec:impossibility}), or it constrains the action space to
actions the model deems safe (a structural approach, restricting
expressiveness to enable governance). In both cases, the two-boundary
analysis applies: the question is whether the governance boundary
(the world model's coverage) matches the expressiveness boundary (the
actions available to the agent).

\paragraph{Formal verification of AI systems.}
Seshia et al.~\cite{seshia2022toward} survey formal methods for
AI. Huang et al.~\cite{huang2020survey} survey verification of
neural networks. These works verify properties of the AI component
itself (robustness, fairness). Our work verifies properties of the
governance architecture that wraps AI components: that governance
is total, that provenance is complete, that the expressiveness and
governance boundaries are identical. The two approaches are
complementary: verifying the governance boundary does not verify
the AI component's behavior, and vice versa.

\paragraph{Levels of abstraction.}
Floridi's levels of abstraction framework~\cite{floridi2008levels}
provides a philosophical foundation for the distinction between content
governance and effect governance. Our analysis operates at the level of
effects: actions that change world state. Content governance (what
models say), intent governance (what agents plan), and systemic
governance (how populations of agents interact) operate at different
levels requiring different frameworks. Our results apply to the effect
level and do not claim to generalize to other levels.

\paragraph{Policy frameworks.}
OPA (Open Policy Agent)~\cite{opa2024} provides policy-as-code for
infrastructure. Rego policies are decidable by construction (no
loops, bounded evaluation). This is a form of structural governance
for infrastructure but does not address the effect-production layer
of AI systems. The structural approach described here operates at a
different level: not ``does this API call match the policy?'' but
``can this component produce API calls at all?''

\section{Conclusion}
\label{sec:conclusion}

The governance problem for agentic AI has a precise structure. Every
effectful system has two boundaries. Those boundaries diverge unless
the architecture makes them identical. Rice's theorem proves the
divergence is undecidable in the general case for any approach that
attempts to decide semantic properties of Turing-complete programs.

The structural alternative operates at a different level. Code
computes. A governed boundary performs effects. The governance
property is syntactic (does this directive have the required
capability?), not semantic (will this program violate the policy?).
Rice does not apply to syntactic properties of structured data.

We propose coterminous governance (Definition~\ref{def:coterminous})
as the testable criterion for AI effect governance. The test is
simple: can you prove that your expressiveness boundary and your
governance boundary are identical? If yes, there is no ungoverned
region and no theater region. If no, both are structurally
inevitable, regardless of how many governance layers are added.

The formal proofs are in the companion
paper~\cite{mccann2026mechanized}: 454 theorems, zero admitted
lemmas, machine-checked in Rocq.\footnote{The Rocq development is available at \url{https://github.com/mashin-live/governance-proofs}.} The Safety Theorem proves that
every expressible program is governed. The Sufficiency Theorem
proves that the governed primitives are Turing-complete. Together,
they prove coterminous governance: the two boundaries are one.

Behavioral approaches (RLHF, content filtering, Constitutional AI)
remain essential for governing model outputs: shaping what AI systems
say. Structural governance addresses a different level: ensuring that
whatever models produce, the system's actions in the world are
governed. The two levels are complementary. At the effect level,
behavioral governance cannot achieve complete coverage even in
principle under arbitrary expressiveness without architectural
totality. This does not mean behavioral governance provides zero
value; it means behavioral governance cannot guarantee that every
effect passes through a governance boundary, regardless of how many
layers are added. This is not a claim about the quality of current
behavioral approaches. It is a consequence of Rice's theorem applied
to the specific problem of effect governance. You cannot govern effects you cannot decide. You
can decide effects you structurally constrain.

\subsection*{Limitations}

\paragraph{Scope limited to effects.}
This paper addresses effect governance: the problem of ensuring that
every action an AI system performs in the world passes through a
governance boundary. Governance of content (what models say), intent
(what agents plan), and systemic behavior (how populations of agents
interact) require different frameworks. We do not claim that structural
governance replaces RLHF, content filtering, or Constitutional AI;
these remain essential for governing model outputs.

\paragraph{Requires re-architecture.}
The analysis assumes a system architecture where computation and effect
can be separated. Systems that mix computation and I/O at the language
level, which includes most existing agent frameworks, would require
re-architecture, not an additional governance layer. Structural
governance cannot be retrofitted onto existing systems as a library or
middleware. It is an architectural commitment that must be made before
the system is built.

\paragraph{Policy correctness not guaranteed.}
The coterminous governance criterion is a structural property, not a
policy correctness guarantee. A system can satisfy coterminous
governance while enforcing a bad policy. Structural governance enforces
whatever policy it is given with mathematical certainty, including
policies that are wrong, incomplete, or harmful. Policy design remains
a human judgment problem. Structural governance is a necessary but not
sufficient condition for good governance.

\paragraph{Restricted sublanguages.}
The Rice's theorem argument applies to the general case: Turing-complete
programs governed by semantic properties. Specific restricted
sublanguages may admit decidable effect properties, though at the cost
of reduced expressiveness. The tradeoff between expressiveness and
decidability is itself a design choice, and for some domains a
restricted sublanguage may be the right one.
Dalrymple et al.~\cite{dalrymple2024guaranteed} identify a similar
structural requirement in their ``guaranteed safe AI'' framework,
where safety must be a property of the system architecture rather than
a behavioral claim.

\paragraph{Practical usability.}
This paper does not address the practical usability implications of
structural governance: developer experience, debugging overhead, the
cost of expressing programs as pure computations plus directives rather
than as imperative code with direct I/O access. These are real
engineering concerns. The separation of computation from effect imposes
a discipline that may slow prototyping and steepen learning curves.
Whether the governance guarantees justify the development cost depends
on the deployment context and the consequences of governance failure.

\paragraph{Companion papers.}
This paper establishes the structural argument and the coterminous governance criterion. Companion papers develop the formal and practical consequences.
\cite{mccann2026mechanized}~mechanizes the safety, invariance, sufficiency, normal form, and necessity results in Rocq (454 theorems, 36 modules, zero admitted). Recent modules extend governance across network boundaries (compositional governance preservation, capability narrowing, protocol uniformity) and through temporal policy evolution (safety under restriction, provenance continuity, rollback safety), strengthening the coterminous claim beyond single-node execution.
\cite{mccann2026gcc}~extends these to seven expressiveness properties, proving that governance preserves semantic transparency and that the four primitives are minimal.
\cite{mccann2026algebraic}~lifts the system-specific results to a parametric algebraic semantics: a three-axiom GovernanceAlgebra that induces a symmetric monoidal category with verified coherence, and extracts the governance kernel to a verified OCaml NIF running in the BEAM runtime.
\cite{mccann2026purity}~discharges the pure module constraint that the safety theorems assume, replacing convention-based enforcement with WASM compilation and cryptographic purity certificates.
\cite{mccann2026provenance}~extends the governance boundary to the supply chain: dual-signature distribution provenance with six-level verification, completing the lifecycle from build through distribution to runtime governance.

\bibliographystyle{plainnat}
\bibliography{two-boundaries-references}

\begin{thebibliography}{33}
\providecommand{\natexlab}[1]{#1}
\providecommand{\url}[1]{\texttt{#1}}
\expandafter\ifx\csname urlstyle\endcsname\relax
  \providecommand{\doi}[1]{doi: #1}\else
  \providecommand{\doi}{doi: \begingroup \urlstyle{rm}\Url}\fi

\bibitem[Bai et~al.(2022)Bai, Kadavath, Kundu, Askell, Kernion, Jones, Chen,
  Goldie, Mirhoseini, McKinnon, et~al.]{bai2022constitutional}
Yuntao Bai, Saurav Kadavath, Sandipan Kundu, Amanda Askell, Jackson Kernion,
  Andy Jones, Anna Chen, Anna Goldie, Azalia Mirhoseini, Cameron McKinnon,
  et~al.
\newblock Constitutional {AI}: Harmlessness from {AI} feedback.
\newblock \emph{arXiv preprint arXiv:2212.08073}, 2022.

\bibitem[Dalrymple et~al.(2024)Dalrymple, Skalse, Bengio, Russell, Tegmark,
  Seshia, Omohundro, Szegedy, Goldhaber, Ammann,
  et~al.]{dalrymple2024guaranteed}
David Dalrymple, Joar Skalse, Yoshua Bengio, Stuart Russell, Max Tegmark,
  Sanjit Seshia, Steve Omohundro, Christian Szegedy, Ben Goldhaber, Nora
  Ammann, et~al.
\newblock Towards guaranteed safe {AI}: A framework for ensuring robust and
  reliable {AI} systems.
\newblock \emph{arXiv preprint arXiv:2405.06624}, 2024.

\bibitem[Dennis and Van~Horn(1966)]{dennis1966programming}
Jack~B. Dennis and Earl~C. Van~Horn.
\newblock Programming semantics for multiprogrammed computations.
\newblock \emph{Communications of the ACM}, 9\penalty0 (3):\penalty0 143--155,
  1966.
\newblock \doi{10.1145/365230.365252}.

\bibitem[Floridi(2008)]{floridi2008levels}
Luciano Floridi.
\newblock The method of levels of abstraction.
\newblock \emph{Minds and Machines}, 18\penalty0 (3):\penalty0 303--329, 2008.

\bibitem[Gifford and Lucassen(1986)]{gifford1986integrating}
David~K. Gifford and John~M. Lucassen.
\newblock Integrating functional and imperative programming.
\newblock In \emph{ACM Conference on LISP and Functional Programming}, pages
  28--38, 1986.
\newblock \doi{10.1145/319838.319848}.

\bibitem[{Guardrails AI}(2024)]{guardrailsai2024}
{Guardrails AI}.
\newblock Guardrails: Adding guardrails to large language models.
\newblock \url{https://github.com/guardrails-ai/guardrails}, 2024.

\bibitem[Haas et~al.(2017)Haas, Rossberg, Schuff, Titzer, Holman, Gohman,
  Wagner, Zakai, and Bastien]{haas2017wasm}
Andreas Haas, Andreas Rossberg, Derek~L. Schuff, Ben~L. Titzer, Michael Holman,
  Dan Gohman, Luke Wagner, Alon Zakai, and J.~F. Bastien.
\newblock Bringing the web up to speed with {WebAssembly}.
\newblock In \emph{ACM SIGPLAN Conference on Programming Language Design and
  Implementation (PLDI)}, pages 185--200, 2017.
\newblock \doi{10.1145/3062341.3062363}.

\bibitem[Huang et~al.(2020)Huang, Kroening, Ruan, Sharp, Sun, Thesing, Wu, and
  Yi]{huang2020survey}
Xiaowei Huang, Daniel Kroening, Wenjie Ruan, James Sharp, Youcheng Sun, Emese
  Thesing, Min Wu, and Xinping Yi.
\newblock A survey of safety and trustworthiness of deep neural networks:
  Verification, testing, adversarial attack and defence, and interpretability.
\newblock \emph{Computer Science Review}, 37:\penalty0 100270, 2020.
\newblock \doi{10.1016/j.cosrev.2020.100270}.

\bibitem[Khattab et~al.(2023)Khattab, Singhvi, Maheshwari, Zhang, Santhanam,
  Vardhamanan, Haq, Sharma, Joshi, Mober, et~al.]{khattab2023dspy}
Omar Khattab, Arnav Singhvi, Paridhi Maheshwari, Zhiyuan Zhang, Keshav
  Santhanam, Sri Vardhamanan, Saiful Haq, Ashutosh Sharma, Thomas~T. Joshi,
  Hanna Mober, et~al.
\newblock {DSPy}: Compiling declarative language model calls into
  self-improving pipelines.
\newblock arXiv preprint arXiv:2310.03714, 2023.

\bibitem[Leijen(2017)]{leijen2017koka}
Daan Leijen.
\newblock Type directed compilation of row-typed algebraic effects.
\newblock \emph{Proceedings of the ACM on Programming Languages}, 1\penalty0
  (POPL):\penalty0 1--28, 2017.
\newblock \doi{10.1145/3009837.3009872}.

\bibitem[Lindley et~al.(2017)Lindley, McBride, and
  McLaughlin]{lindley2017frank}
Sam Lindley, Conor McBride, and Craig McLaughlin.
\newblock Do be do be do.
\newblock In \emph{Proceedings of the ACM on Programming Languages (POPL)},
  pages 1--26, 2017.
\newblock \doi{10.1145/3009837.3009897}.

\bibitem[Lud{\"a}scher et~al.(2006)Lud{\"a}scher, Altintas, Berkley, Higgins,
  Jaeger, Jones, Lee, Tao, and Zhao]{ludaescher2006kepler}
Bertram Lud{\"a}scher, Ilkay Altintas, Chad Berkley, Dan Higgins, Efrat Jaeger,
  Matthew Jones, Edward~A. Lee, Jing Tao, and Yang Zhao.
\newblock Scientific workflow management and the {Kepler} system.
\newblock \emph{Concurrency and Computation: Practice and Experience},
  18\penalty0 (10):\penalty0 1039--1065, 2006.
\newblock \doi{10.1002/cpe.994}.

\bibitem[McCann(2026{\natexlab{a}})]{mccann2026algebraic}
Alan~L. McCann.
\newblock Algebraic semantics of governed execution: Monoidal categories,
  effect algebras, and coterminous boundaries, 2026{\natexlab{a}}.

\bibitem[McCann(2026{\natexlab{b}})]{mccann2026gcc}
Alan~L. McCann.
\newblock Effect-transparent governance for {AI} workflow architectures:
  Semantic preservation, expressive minimality, and decidability boundaries,
  2026{\natexlab{b}}.

\bibitem[McCann(2026{\natexlab{c}})]{mccann2026mechanized}
Alan~L. McCann.
\newblock Mechanized foundations of structural governance: Machine-checked
  proofs for governed intelligence, 2026{\natexlab{c}}.

\bibitem[McCann(2026{\natexlab{d}})]{mccann2026provenance}
Alan~L. McCann.
\newblock Cryptographic registry provenance: Structural defense against
  dependency confusion in {AI} package ecosystems, 2026{\natexlab{d}}.
\newblock arXiv preprint (submitted).

\bibitem[McCann(2026{\natexlab{e}})]{mccann2026purity}
Alan~L. McCann.
\newblock Certified purity for cognitive workflow executors: From static
  analysis to cryptographic attestation, 2026{\natexlab{e}}.

\bibitem[Miller(2006)]{miller2006robust}
Mark~S. Miller.
\newblock \emph{Robust Composition: Towards a Unified Approach to Access
  Control and Concurrency Control}.
\newblock PhD thesis, Johns Hopkins University, 2006.

\bibitem[Moggi(1991)]{moggi1991notions}
Eugenio Moggi.
\newblock Notions of computation and monads.
\newblock \emph{Information and Computation}, 93\penalty0 (1):\penalty0 55--92,
  1991.
\newblock \doi{10.1016/0890-5401(91)90052-4}.

\bibitem[Myers(1999)]{myers2001jif}
Andrew~C. Myers.
\newblock {JFlow}: Practical mostly-static information flow control.
\newblock In \emph{ACM Symposium on Principles of Programming Languages
  (POPL)}, pages 228--241, 1999.
\newblock \doi{10.1145/292540.292561}.

\bibitem[Myers and Liskov(1997)]{myers1997decentralized}
Andrew~C. Myers and Barbara Liskov.
\newblock A decentralized model for information flow control.
\newblock In \emph{ACM Symposium on Operating Systems Principles (SOSP)}, pages
  129--142, 1997.
\newblock \doi{10.1145/268998.266669}.

\bibitem[{Open Policy Agent}(2024)]{opa2024}
{Open Policy Agent}.
\newblock {OPA}: Open policy agent.
\newblock \url{https://www.openpolicyagent.org/}, 2024.

\bibitem[Ouyang et~al.(2022)Ouyang, Wu, Jiang, Almeida, Wainwright, Mishkin,
  Zhang, Agarwal, Slama, Ray, et~al.]{ouyang2022training}
Long Ouyang, Jeff Wu, Xu~Jiang, Diogo Almeida, Carroll~L. Wainwright, Pamela
  Mishkin, Chong Zhang, Sandhini Agarwal, Katarina Slama, Alex Ray, et~al.
\newblock Training language models to follow instructions with human feedback.
\newblock \emph{Advances in Neural Information Processing Systems},
  35:\penalty0 27730--27744, 2022.

\bibitem[Peyton~Jones(2001)]{peytonjones2001tackling}
Simon Peyton~Jones.
\newblock Tackling the awkward squad: Monadic input/output, concurrency,
  exceptions, and foreign-language calls in {Haskell}.
\newblock In \emph{Engineering Theories of Software Construction}, pages
  47--96. IOS Press, 2001.

\bibitem[Plotkin and Pretnar(2009)]{plotkin2009handlers}
Gordon Plotkin and Matija Pretnar.
\newblock Handlers of algebraic effects.
\newblock In \emph{European Symposium on Programming (ESOP)}, pages 80--94,
  2009.
\newblock \doi{10.1007/978-3-642-00590-9_7}.

\bibitem[Rebedea et~al.(2023)Rebedea, Dinu, Sreedhar, Parisien, and
  Cohen]{rebedea2023nemo}
Traian Rebedea, Razvan Dinu, Makesh~Narsimhan Sreedhar, Christopher Parisien,
  and Jonathan Cohen.
\newblock {NeMo} guardrails: A toolkit for controllable and safe {LLM}
  applications with programmable rails.
\newblock In \emph{Conference on Empirical Methods in Natural Language
  Processing (EMNLP), System Demonstrations}, 2023.

\bibitem[Rice(1953)]{rice1953classes}
Henry~Gordon Rice.
\newblock Classes of recursively enumerable sets and their decision problems.
\newblock \emph{Transactions of the American Mathematical Society}, 74\penalty0
  (2):\penalty0 358--366, 1953.

\bibitem[Seshia et~al.(2022)Seshia, Sadigh, and Sastry]{seshia2022toward}
Sanjit~A. Seshia, Dorsa Sadigh, and S.~Shankar Sastry.
\newblock Toward verified artificial intelligence.
\newblock \emph{Communications of the ACM}, 65\penalty0 (7):\penalty0 46--55,
  2022.
\newblock \doi{10.1145/3503914}.

\bibitem[Stefan et~al.(2011)Stefan, Russo, Mitchell, and
  Mazi{\`e}res]{stefan2011lio}
Deian Stefan, Alejandro Russo, John~C. Mitchell, and David Mazi{\`e}res.
\newblock Flexible dynamic information flow control in {Haskell}.
\newblock In \emph{ACM SIGPLAN Haskell Symposium}, pages 95--106, 2011.
\newblock \doi{10.1145/2034675.2034688}.

\bibitem[Wadler(1995)]{wadler1995monads}
Philip Wadler.
\newblock Monads for functional programming.
\newblock In \emph{Advanced Functional Programming}, volume 925 of \emph{LNCS},
  pages 24--52. Springer, 1995.
\newblock \doi{10.1007/3-540-59451-5_2}.

\bibitem[Wahbe et~al.(1993)Wahbe, Lucco, Anderson, and Graham]{wahbe1993sfi}
Robert Wahbe, Steven Lucco, Thomas~E. Anderson, and Susan~L. Graham.
\newblock Efficient software-based fault isolation.
\newblock In \emph{ACM Symposium on Operating Systems Principles (SOSP)}, pages
  203--216, 1993.
\newblock \doi{10.1145/168619.168635}.

\bibitem[Wei et~al.(2023)Wei, Haghtalab, and Steinhardt]{wei2023jailbroken}
Alexander Wei, Nika Haghtalab, and Jacob Steinhardt.
\newblock Jailbroken: How does {LLM} safety training fail?
\newblock In \emph{Advances in Neural Information Processing Systems},
  volume~36, 2023.

\bibitem[Zou et~al.(2023)Zou, Wang, Kolter, and Fredrikson]{zou2023universal}
Andy Zou, Zifan Wang, J.~Zico Kolter, and Matt Fredrikson.
\newblock Universal and transferable adversarial attacks on aligned language
  models.
\newblock \emph{arXiv preprint arXiv:2307.15043}, 2023.

\end{thebibliography}

\end{document}